# Traffic Congestion Prediction using Deep Convolutional Neural Networks: A Color-coding Approach

Mirza Fuad Adnan
*Department of Electrical and Electronic Engineering*
*Islamic University of Technology*
Gazipur, Dhaka, Bangladesh
adnan16@iut-dhaka.edu

Nadim Ahmed
*Department of Electrical and Electronic Engineering*
*Islamic University of Technology*
Gazipur, Dhaka, Bangladesh
nadimahmed@iut-dhaka.edu

Imrez Ishraque
*Department of Electrical and Electronic Engineering*
*Islamic University of Technology*
Gazipur, Dhaka, Bangladesh
imrezishraque@iut-dhaka.edu

Md. Sifath Al Amin
*Department of Electrical and Electronic Engineering*
*Islamic University of Technology*
Gazipur, Dhaka, Bangladesh
sifath19@iut-dhaka.edu

Md. Sumit Hasan
*Department of Electrical and Electronic Engineering*
*Islamic University of Technology*
Gazipur, Dhaka, Bangladesh
sumithasan@iut-dhaka.edu

*Abstract*— The traffic video data has become a critical factor in confining the state of traffic congestion due to the recent advancements in computer vision. This work proposes a unique technique for traffic video classification using a color-coding scheme before training the traffic data in a Deep convolutional neural network. At first, the video data is transformed into an imagery data set; then, the vehicle detection is performed using the You Only Look Once algorithm. A color-coded scheme has been adopted to transform the imagery dataset into a binary image dataset. These binary images are fed to a Deep Convolutional Neural Network. Using the UCSD dataset, we have obtained a classification accuracy of 98.2%.

## I. INTRODUCTION

Urbanization's acceleration speeds up traffic problems, resulting in economic losses and immobilization of urban functions [1]. The effects of traffic congestion extend to the individual as well. Some significant effects of traffic congestion are the abundant waste of time, particularly during peak hours, mental fatigue, and additional pollution, contributing to catastrophic natural outcomes. The cost of traffic congestion in the four nations of France, Germany, the United Kingdom, and the United States is expected to increase by 55 billion US dollars by 2030 [2]. A nation cannot progress without ensuring economic growth and the comfort of its road users, which is impossible without efficient traffic flow. The ability to foresee traffic congestion gives officials and consumers the necessary time to allocate resources to ensure that travelers' journeys go smoothly. Consequently, a generic and widely applicable traffic congestion-detecting system is currently required [3].

Modern Information and Communication Technologies and the Internet of Things have contributed to the development of Intelligent Transportation Systems (ITSs), which have enabled the application of Traffic Forecasting (TF) techniques [4]. Starting with a theoretical perspective and progressing to data-driven methodologies, an abundance of research has been undertaken on traffic congestion forecasting and relevant topics. Traditionally spot-based sensors are used for traffic estimation. Whenever a vehicle stops over a loop or goes by the loop, the sensor will count the number of vehicles passed. Inductive loop sensors, piezoelectric sensors, and magnetic loops are widely used technologies in the traffic estimation ecosystem [5], But these sensors are costly, which makes them hard to implement on a wide scale. Also, these spot sensors can only quantify or measure traffic flow or address certain subtasks (e.g., traffic queue measurement and traffic density detection). As a result, this cannot be a sustainable generic model to apply [6]. On top of that, recent developments in infrared and laser radar sensors have prompted the gradual replacement of conventional spot-based sensors with most of these devices [5]. Apart from these sensor-based technologies, a more affordable method of gathering network-wide traffic data is emerging, utilizing Global Position System (GPS) devices. Through tracking vehicle trajectories, in-vehicle GPS technology enables recording vehicle speed and location at a given time. It allows them to follow vehicle trajectories and evaluate traffic condition performance in a broad network at a reasonable cost, making them popular in large-scale research. However, this approach has certain disadvantages; firstly, speed is the only considered parameter, which can sometimes lead to the loss of confidential information from unprocessed GPS data and an inaccurate assessment of traffic congestion [7]. Secondly, as GPS-based intelligent traffic systems depend on the number of vehicles employing GPS, the detection precision of traffic updates decreases drastically when the number of GPS-equipped vehicles starts to decline significantly [1]. Additionally, Applying GPS-based data to detect traffic congestion on arterial roads will become a complicated scenario because of manual manipulations [7]. Intelligent traffic monitoring system includes various surveillance equipment and video monitoring to calculate real-time traffic data. These systems effectively monitor traffic scenarios and activities, maintain records, gather evidence, and arrive at the right decisions. Nevertheless, these systems are not well suited for intelligent traffic management due to limited prediction capabilities, high labor, and maintenance costs, and vulnerable privacy.

On the other hand, TF has shifted from a traffic theory-based approach to a data-driven standpoint due to the increased diversity and number of available traffic data provided by ITSs [4]. Vision-based detection technologies have improved significantly in recent years. Advanced image processing algorithms and object detection techniques have created a new opportunity for vision-based intelligent traffic management systems [8]. Different statistical forecasting approaches, shallow machine learning algorithms, and deep learning methods have achieved impressive precision in these classification tasks [3]. Principal benefits of these vision-based approaches:

- They do not rely on picking features manually. Consequently, this eliminates the restrictions on systems utilizing camera images for traffic-state evaluation and forecasting [5].

- Short-term to long-term estimation of traffic congestion (starting from a few minutes to even for multiple hours).

- This detection system can automatically communicate with other entities of a traffic network, thereby contributing to a more optimal environment for traffic management [9].

- More information and parameters about traffic can be considered to increase the system's efficacy [3].

- A higher number of generic data solutions with a minimized cost and efficient performance [3].

- Easy maintenance and can continuously be updated and driven into a more authentic version with relevant sets of recent data [1].

Research has been conducted to estimate traffic situations by fusing multiple data sources for traffic state estimation with visual inputs integrated with machine learning and deep learning approaches. Akhtar and Moridpour [3] shows a direct comparison between Shallow machine learning (SML) algorithms and Deep Machine Learning (DML) algorithms is shown here analyzing different several notable research works. Most of these works evaluated five parameters: traffic occupancy, congestion volume, and vehicle density, along with traffic congestion index and total traveling time while predicting and analyzing the overall traffic occupancy. After analyzing these works, it is visible that deep learning algorithms can assess a large dataset more efficiently. As a result, they proved it to be more effective than Shallow machine learning (SML) methods like artificial neural network (ANN) and support vector machine (SVM) in this field. P. Chakraborty et al.[5] used two deep neural networks: deep convolutional neural networks (DCNNs) and you only look once (YOLO) and compared these results with support vector machine (SVM) model to analyze the advantages of deep learning models. In order to produce both short-term and long-term forecasts(from 5 minutes to up to 4 hours), T. Bogaerts et al. [4] built a hybrid Deep Neural Network that concurrently extracts the spatial aspects of traffic using graph convolution and its temporal features using Long Short Term Memory (LSTM) cells. In addition, they selected the most appropriate road linkages for both short- and long-term TF using a data reduction technique, which increased their efficiency. M.A.A Al-qaness et al. [8] demonstrated an intelligent video surveillance-based vehicle tracking system that can recognize, track, and count vehicles in various situations by combining neural networks, image-based tracking, and You Only Look Once (YOLOv3). H.Cui et al. [9] employ two convolutional networks, AlexNet and GoogLeNet, to characterize traffic congestion situations. The images in the dataset here were taken from traffic surveillance images. In spite of having obscured visual features in some pictures in the dataset, AlexNet and GoogLeNet showed a convincing result by successfully classifying the images and recognizing highway traffic congestion. In [1], leveraging a gray-level co-occurrence matrix, the multi-layered detection method first determines the density of surrounding items. Then the velocity of moving objects is then determined by incorporating the Lucas–Kanade optical flow along with pyramid implementation. Then Gaussian mixture model is applied to demonstrate the model further tuned by CNN. Crc3d is a suggested mapping in [7] to cube framework to forecast the urban traffic pattern for the holistic network utilizing 3-dimensional convolution networks, convolutional neural networks(CNN), and recurrent neural networks (RNN). The architecture incorporates spatial and temporal dimensions features combining C3D and CNN-RNN. H. Nguyen [10] suggests an enhanced vehicle detection scheme based on an accelerated R-CNN. It used the mobile net architecture to build the convolution layer. A soft NMS algorithm was deployed to address the problem of redundant proposals. Context-aware ROI pooling layer was used to scale the suggestions to the required dimensions, and MobileNet architecture was used for constructing the classifier as the method's final step. A hybrid 2D-3D CNN model-based driving assistance system was developed in [11], which uses a transfer learning paradigm. In the first architecture, hybrid-TSR is designed to perform the duty of traffic sign recognition efficiently. The second framework, called hybrid-SRD, enables semantic road space classification by a blending of up-sampling and deconvolutional methods. D. Impedovo et al. [12] used feature extraction to classify the congestions. Various objection detection algorithms were used for vehicle detection and feature extraction which were then compared side by side by applying machine learning classifiers and deep learning methods to see which better gave output. It is visible in [13] that ITSC or intelligent traffic signal control can be used to solve the traffic congestion problem efficiently. Here Reinforcement Learning was applied to vehicles detected by ITSC system to reduce average waiting time. R. Cucchiara et al. [6] propose a 2-level traffic monitoring system called VTTS (Vehicular traffic tracking system) based on vehicle detection and tracking. Vehicle detection is done using Dedicated Short-Range Communication (DRSC) structures in intersections rather than by a camera or loop detector. The reduced image processing components harvest visual data under different situations, while the elevated units track the vehicles.

In this paper, we are interested in investigating a new color coding-based scheme to train a Deep Convolutional Neural Network (DCNN) and evaluate its performance in forecasting Traffic congestion in sites that differ significantly from the training data set. The general outline of our proposed scheme is illustrated in Fig. 1.

We have used the UCSD highway traffic data to train our proposed DCNN. A binary imagery dataset of highway traffic was built and labeled with around 3000 congestion and 1000 non-congestion images. The performance of DCNN has been investigated using the binary image dataset. DCNN achieved 98.2% of classification accuracy. We also validate our scheme by feeding test images of diverse sites; almost 97% of the cases have been precisely recognized. Our proposed color-coding scheme can drive this congestion challenge more independent of the dataset.

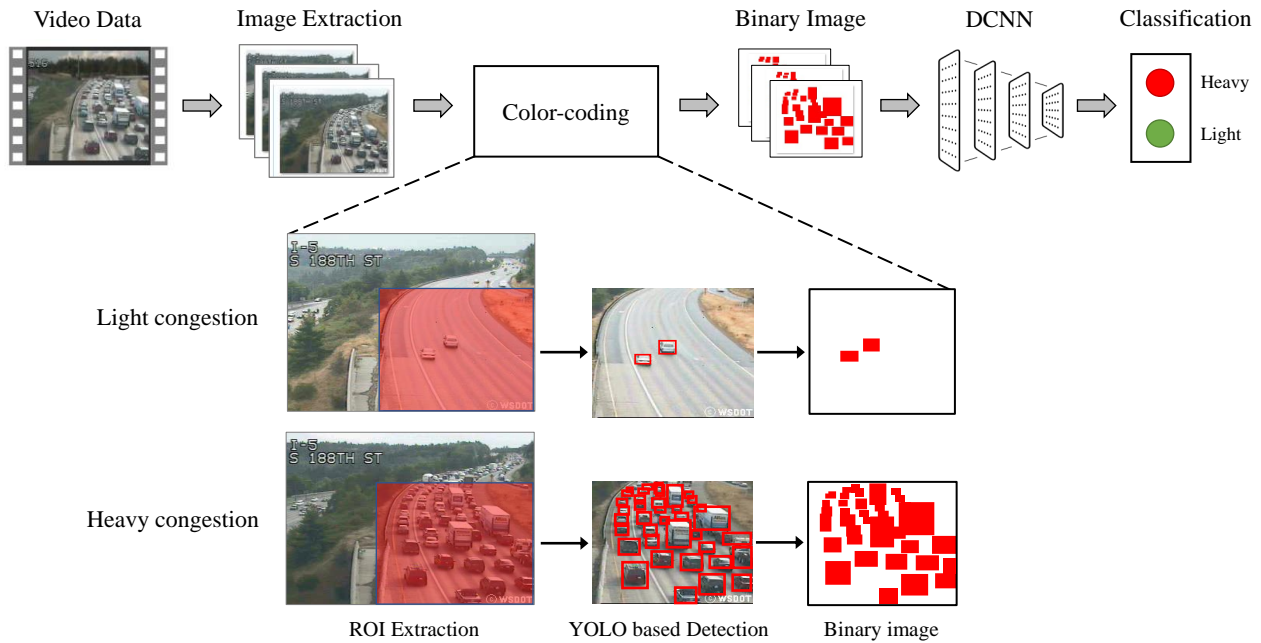

Fig. 1. Framework of color-coding based Traffic congestion Detection

## II. IMAGE DATA

The UCSD dataset contains 254 daytime highway traffic videos. A stationary camera recorded these videos.

Since this paper focuses on processing still images, images are extracted from these videos. The dataset comprises diverse traffic patterns like light, medium, and heavy congestion. Hand-labeled ground truth has been provided, which describes each video sequence. Table exhibits a summary of the UCSD dataset.

The medium type tends to be selected for marginal cases in which it is tough to comment on whether the traffic is congested or not. Hence, the medium congestion types are rejected from use. Each image is labeled by two different annotators (congested or non-congested). After dismissing the medium types, we now have 4110 labeled examples, 2990 uncongested images, and 1120 congested images.

## III. METHODOLOGY

The traffic congestion Detection framework followed in this study includes three main steps.

### A. Vehicle Detection

Vehicle Detection is the most crucial step in this congestion detection framework. The purpose is to portray the number of vehicles in the road segment accurately. For accurate detection of the vehicles, several techniques have been proposed, like Frame differencing [14], Optical flow [15], Region-based Convolutional Neural Network (R-CNN) [16], and You Only Look Once (YOLO) [17]. While all the previous vehicle detection algorithms employ regions to identify the vehicle within the image, YOLO uses regression, where the classes and bounding boxes for the whole image are predicted in a single run. Hence, YOLO is considered one of the fastest vehicle detection techniques because it requires only one image processing. However, accuracy decreases when two different Vehicles are close to each other. In this study, we have adopted a YOLOv3 object detector consisting of a CNN called Darknet, where 24 convolutional layers work as feature extractors and two dense layers for prediction.

### B. Road Congestion Estimation

Traffic congestion occurs when the volume of road traffic is so high that it can ultimately propel the motorists to slow down or stop altogether. Here we took the UCSD Highway traffic dataset, a state-of-the-art vehicular traffic state classification dataset for traffic congestion prediction [18]. Upon the accurate detection of vehicles from the traffic scene, a color coding scheme has been proposed in this study. We developed a technique by segmenting each vehicle-detected bounding box by red color, keeping the undetected region white. Hence, the image containing multiple vehicles has been converted into a binary image of red and white pixels, where the red pixel indicates the possibilities of a vehicle and the white pixel suggests the possibilities of occupancy. Now our dataset has been boosted to a more noteworthy interpretation. Earlier it was merely an image containing concealed vehicles. Visual features [12] (total number of vehicles, Traffic velocity, Traffic flow, etc.) could be used to forecast traffic congestion alongside deep learning techniques. However, raw images are fed to these networks, making the deep neural network model more dependent on the dataset. Hence, our color-coded images Showed in Fig. 2 and Fig. 3 are more independent of the training dataset as it transforms any scene into a binary image.

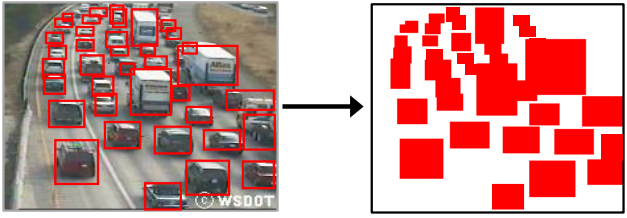

Fig. 2. Heavy Congestion Transformation to Binary image

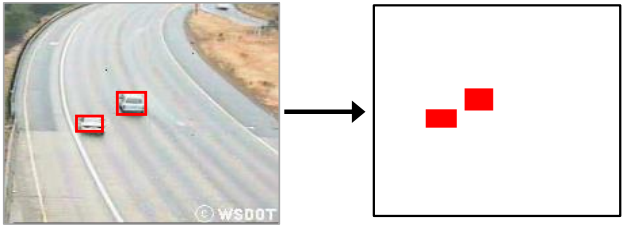

Fig. 3. Light Congestion Transformation to Binary image

## C. Classification

Deep convolution neural networks (DCNNs) are regarded as the state-of-the-art image classification approach. A traditional ConvNet architecture has been used in this study consisting of convolution layers and pooling layers. The UCSD data set contains 254 highway traffic videos. Each video has 40-50 frames (320x240 resolution). 4-5 frames from each video have been extracted, and then a region of interest (ROI) is set manually to minimize the scenes' unrelated external objects. The trimmed images were then resized to 180x180 pixels to avert memory allocation problems during the training of the model. These images were fed into the model as two successive convolution layers 32x3x3 in size, followed by a max pooling layer 2x2 in size. Then again, the network was extended by two more convolution layers of 64x3x3 followed by another max pooling layer of the same size. A dropout layer of 0.25 probability was assigned after each max pooling layer to prevent overfitting. Throughout the model, ReLU was used as an activation function. Table.1 Shows the DCNN model architecture.

TABLE I. DCNN MODEL ARCHITECTURE USED

| Layer | Kernel | Stride | Output shape |
|---|---|---|---|
| Input | | | [180, 180, 3] |
| Convolution | 3x3 | 1 | [180,180, 32] |
| Convolution | 3x3 | 1 | [178, 178, 32] |
| Max Pooling | 2x2 | 2 | [89, 89, 32] |
| Dropout | | | [89, 89, 64] |
| Convolution | 3x3 | 1 | [89, 89, 64] |
| Convolution | 3x3 | 1 | [87, 87, 64] |
| Max Pooling | 2x2 | 2 | [43, 43, 64] |
| Dropout | | | [43, 43, 64] |
| Dense | | | 512 |
| Dropout | | | 512 |
| Dense | | | 2 |

DCNN, being computationally expensive, requires an immense number of images to train the model to handle overfitting. However, in this study, we ran out of data. A total of 1170 images have been generated from the 195 videos by frame extraction. So, around 1200 images were available for training. Data Augmentation technique [19] alongside dropout regularization has been adapted to prevent overfitting. To ensure data augmentation, sample images were flipped horizontally illustrated in Fig. 5 and vertically illustrated Fig. 4 randomly. It took 30 minutes to train the model on a NVIDIA Quadro M1000M GPU with 8GB RAM memory.

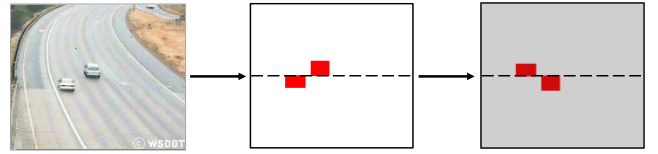

Fig. 4. Data Augmentation (Vertical Flip)

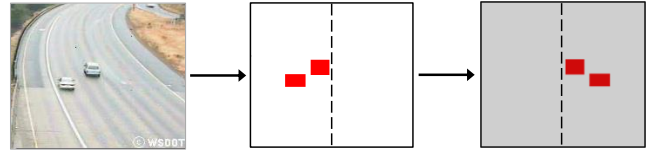

Fig. 5. Data Augmentation (Horizontal Flip)

## IV. RESULTS

We achieved 98.2% classification accuracy through our proposed color-coded scheme. We have employed two Deep Learning Models, one (YOLO) for vehicle detection and a DCNN for classification. The model's performance was evaluated using traditional performance metrics of precision, recall, and accuracy. To assess these metrics, we need to define the concept of true positive (TP), true negative (TN), false positive (FP), and false negative (FN).

**TP**: When a congested image was correctly labeled
**TN**: If a non-congested image was correctly labeled
**FP**: If a congested image was predicted as non-congested
**FN**: If a non-congested image was predicted as congested

Now, Precision, Recall and Accuracy can be found using following equations:

$$Precision = \frac{TP}{TP + FP} \quad (1)$$

$$Recall = \frac{TP}{TP + FN} \quad (2)$$

$$Accuracy = \frac{TP}{TP + FP + TN + FN} \quad (3)$$

The standard metrics of the proposed model is shown in Table II.

TABLE II. STANDARD METRICS

| Performance Metrics | Result (%) |
|---|---|
| Precision | 98.2 |
| Recall | 95.6 |
| Accuracy | 98.2 |

Furthermore, we have validated our model with test images of eclectic sites without pre-training on that site. Our model can perform congestion prediction with a different road segment or location. To validate this, we feed the ROI images before applying the color coding to the DCNN.

Without color coding, the model failed to detect the validation set properly (10% accuracy only).

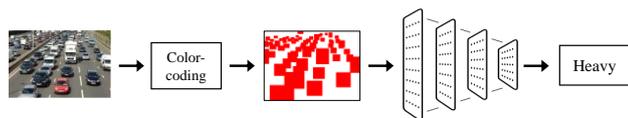

Fig. 6. Validation set using color-coded scheme (correct prediction)

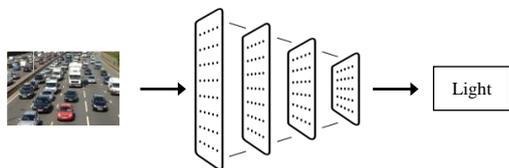

Fig. 7. Validation set without color-coded scheme (false prediction)

From Fig. 6 and Fig. 7 it is clearly evident that color coded scheme enhances the performance of DCNN in terms of data dependency. The validation set is a collection of images of diverse traffic congestion. In Fig. 7 as the input image is from the different road segment by which the DCNN has been trained, it failed to provide the correct classification. On the contrary in Fig. 6 due to the color-coded scheme the prediction is accurate.

## V. CONCLUSION

This paper has proposed a system that uses a color-coding scheme for training DCNN, which would predict traffic congestion with an accuracy of 98.2%. In the previous works, almost all the approaches heavily depend on the dataset [2-5]. All these datasets incorporate different road segments, and diversity in the scenes has also been observed. Then, our color coding of the vehicles would lead us to a more data-independent segmentation. In the present work, these color-coded binary images provide a relationship between occupied space in the road segment by vehicles(red color) with vacant space(white color). Hence, it becomes the more diverse model. An approach could have been taken to accumulate all available traffic congestion Data by using our proposed color-coding scheme to obtain a model which could be used as a benchmark for the traffic congestion classification challenge. Several State-Of-The-Art models like Alexnet, Googlenet, and VGGnet could be explored here for better detection.